\documentclass[final]{cvpr}

\usepackage{import}
\usepackage{times}
\usepackage{epsfig}
\usepackage{graphicx}
\usepackage{amsmath}
\usepackage{amssymb}
\usepackage[toc,page]{appendix}

\usepackage[pagebackref=true,breaklinks=true,colorlinks,bookmarks=false]{hyperref}

\def\cvprPaperID{4259} % *** Enter the CVPR Paper ID here
\def\confYear{CVPR 2021}

\begin{document}

\title{Affective Processes: stochastic modelling of temporal context \\ for emotion and facial expression recognition}

\author{Enrique Sanchez$^1$ \hspace{4pt} Mani Kumar Tellamekala$^2$ \hspace{4pt} Michel Valstar$^2$ \hspace{4pt} Georgios Tzimiropoulos$^{1,3}$ \\
$^1$\,Samsung AI Center, Cambridge, UK\\
$^2$\,University of Nottingham, Nottingham, UK\\
$^3$\,Queen Mary University London, London, UK\\
{\tt\small e.lozano@samsung.com} \hspace{2pt} {\tt\small \{psxmkt, michel.valstar\}@nottingham.ac.uk} \hspace{2pt} {\tt\small g.tzimiropoulos@qmul.ac.uk} 
}

\maketitle

%%%%%%%%% ABSTRACT
\begin{abstract}
   
\noindent Temporal context is key to the recognition of expressions of emotion. Existing methods, that rely on recurrent or self-attention models to enforce temporal consistency, work on the feature level, ignoring the task-specific temporal dependencies, and fail to model context uncertainty. To alleviate these issues, we build upon the framework of Neural Processes to propose a method for apparent emotion recognition with three key novel components: (a) probabilistic contextual representation with a \textit{global} latent variable model; (b) temporal context modelling using task-specific predictions in addition to features;  and (c) smart temporal context selection. We validate our approach on four databases, two for Valence and Arousal estimation (SEWA and AffWild2), and two for Action Unit intensity estimation (DISFA and BP4D). Results show a consistent improvement over a series of strong baselines as well as over state-of-the-art methods.

\end{abstract}

\begin{figure}[htp!]
    \centering
    \includegraphics[width=0.45\textwidth]{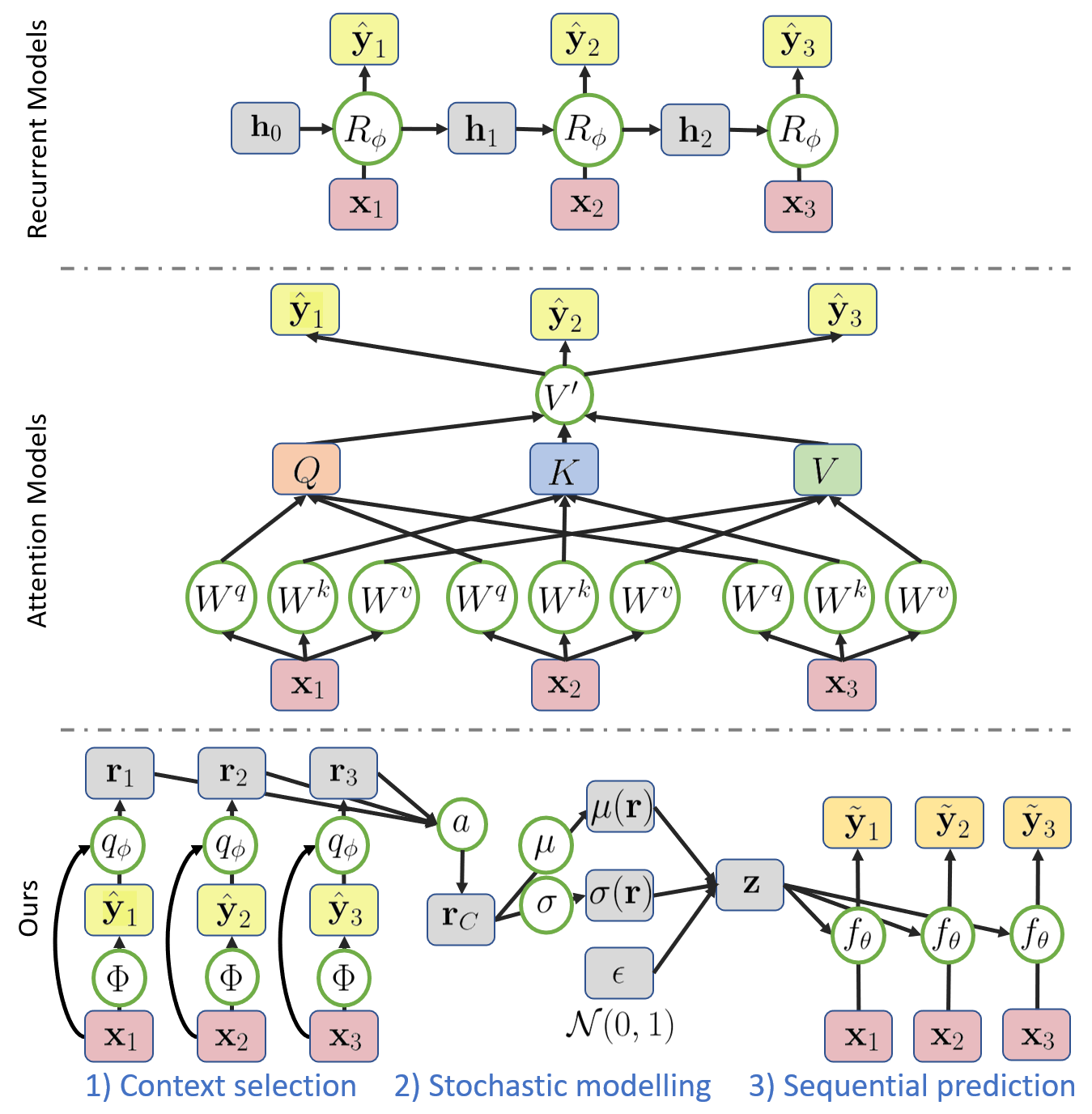}
\caption{Different ways of modelling temporal context in the literature of emotion recognition: Recurrent Models (\textbf{top}), Attention-based Models (\textbf{middle}), and our proposed approach based on Neural Processes (\textbf{bottom}). Our method proposes a probabilistic modelling of temporal context through a global latent variable ${\bf z} \sim \mathcal{N}(\mu, \sigma)$, and includes a novel method for automatic context selection. The global latent variable allows to sample temporal functions $f_\theta(.,{\bf z})$ that are consistent with the captured context.} \vspace{-4pt}
    \label{fig:abstract}
\end{figure}

\section{Introduction}

\noindent In this paper, we address the problem of facial behaviour recognition from video, in particular, the problem of recognising apparent emotions in terms of Valence and Arousal~\cite{posner2005circumplex}, and facial expressions in terms of Action Unit intensity~\cite{ekman1977facial}. This is a longstanding problem in video recognition which has been extensively studied by the computer vision community~\cite{kollias2017recognition, valstar2008timing, martinez2017automatic, jaiswal2016deep, Kossaifi_2020_CVPR, kollias2019deep, walecki2017deep, zafeiriou2017aff, zhang2019context, wang2017ast, sandbach2013markov, rudovic2012kernel, rudovic2012multi, linh2017deepcoder, ntinou2021}. Nevertheless, even recent methods~\cite{kollias2020analysing, Kossaifi_2020_CVPR, zhang2019context} struggle to achieve high accuracy  on the most difficult datasets including SEWA~\cite{kossaifi2019sewa}, Aff-Wild2~\cite{kollias2018aff}, BP4D~\cite{zhang2014bp4d} and DISFA~\cite{mavadati2013disfa}. Hence, there is still great progress to be made towards solving this challenging problem. In this paper, we show that  effective temporal context modelling is a key feature for significantly advancing the state-of-the-art.

But what makes emotion (and facial expression) recognition  such a difficult problem? There are many, often inter-related, reasons for this including: (a) annotation of emotions is often both subjective and laborious making it hard to annotate it consistently \cite{sethu2019ambiguous}; (b) there exist only small- and medium-size emotion video datasets and, due to point (a) above, the annotations for these datasets are often noisy; (c) emotions are subtle, acting as unobserved, latent variables that only partially explain facial appearance over time \cite{barrett2019emotional}. They often require temporal as well as multi-modal context in order to be robustly recognised. From these challenges, this paper focuses on solving problem (c). In particular, it  proposes a completely unexplored perspective for effective incorporation of temporal context into emotion recognition. 

Previous work in temporal modelling of emotions has primarily focused on modelling facial expression dynamics, i.e., the way that facial expressions evolve over time for recognition purposes. A typical deep learning pipeline for this has been a Convolution Neural Network (CNN) followed by a Recurrent Neural Network (RNN), usually an LSTM or GRU~\cite{kollias2020exploiting, kollias2019deep, kollias2017recognition, wang2017ast}. Although these dynamics can be particularly useful in recognising specific facial expressions (e.g.~\cite{valstar2006fully, valstar2008timing, jaiswal2016deep, martinez2017automatic, wang2019novel}), we believe that their importance has been over-emphasised for the problem of emotion recognition.

We posit that temporal context is more important than facial dynamics for emotion recognition. The cues for inferring a person's apparent emotion from their facial behaviour are often sparsely and non-regularly distributed over a temporal window of interest and collecting such distributed contextual cues is critical to inferring emotions robustly. In addition, due to the person-specific variability of facial expressions but, most importantly, due to their subjective annotation, context must be modelled in a stochastic manner. 

To address the aforementioned challenges, in this work, and for the first time to the best of our knowledge, we build upon the framework of Neural Processes~\cite{garnelo2018conditional, garnelo2018neural} to propose a model for emotion recognition with 3 key components: (1) stochastic contextual representation with a \textit{global} latent variable model; (2) task-aware temporal context modelling not only by using features but also task-specific predictions; and (3) effective temporal context selection. 

Fig.~\ref{fig:abstract} depicts an overview of the working flow our method (bottom), RNN-based methods (top), and methods based on self-attention~\cite{vaswani2017attention} (middle). 
Note methods for context modelling based on self-attention do not satisfy any of (1)-(3): They are deterministic and model long-term dependencies not globally, but by using pair-wise feature similarities. Moreover, they do not use task-aware context modelling and do not perform context selection. 

Overall, \textbf{we make} the following \textbf{3 contributions}: \vspace{-2pt}
\begin{enumerate}
    \item 
    We propose Affective Processes: the very first model for emotion recognition with three key properties: (a) global stochastic contextual representation; (b) task-aware temporal context modelling; and (c)  temporal context selection. We conduct a large number of ablation studies illustrating the contribution of each of our model's key features and components.
    \item
    We show that our model is more effective in modelling temporal context not only than CNNs+RNNs but also than a strong baseline based on self-attention. Our model outperforms both baselines by large margin.
    \item
    We validated our approach on the most difficult databases for emotion recognition in terms of Valence and Arousal estimation, namely SEWA~\cite{kossaifi2019sewa} and AffWild2~\cite{kollias2018aff}. We further show that our approach can be effective for the problem of Action Unit intensity estimation on DISFA~\cite{mavadati2013disfa} and BP4D~\cite{zhang2014bp4d} datasets. On all datasets used, we show a consistent and often significant improvement over state-of-the-art methods.
\end{enumerate}

\section{Related Work}

\noindent This section reviews related work from temporal apparent emotion recognition, Action Unit intensity estimation and neural modelling of Stochastic Processes.

\noindent \textbf{Temporal Modelling in Emotion Recognition:} Most existing methods model the  temporal dynamics of continuous emotions  using deterministic approaches such as Time-Delay Neural Networks~\cite{meng2015time}, RNNs, LSTMs and GRUs~\cite{kollias2020exploiting, kollias2019deep, kollias2017recognition, wang2017ast, deng2020multitask, tellamekala2019temporally}, multi-head attention models~\cite{huang2019efficient}, 3D Convolutions~\cite{zhang2020m3f, kuhnke2020two}, 3D ConvLSTMs~\cite{huang2018end}, and temporal-hourglass CNNs~\cite{du2019spatio}. While these deterministic models are capable of effectively learning the temporal dynamics, they do not take the inherent stochastic nature of the continuous emotion labels into account. Deviating from this trend, recently,~\cite{ong2019modeling} applied variational RNNs~\cite{chung2015recurrent} to the valence estimation problem to add stochasticity to the temporal model of emotion recognition. Similarly, in~\cite{oveneke2017leveraging}, emotion is treated as a dynamical latent system state which is estimated using Gaussian processes coupled with Kalman filters, while Deep Extended Kalman filters are used in~\cite{oveneke2019leveraging}. These probabilistic methods have better latent structure-discovery capabilities than the the deterministic temporal models, however, they focus only on the temporal dynamics of emotions, ignoring the temporal context. In contrast, our method primarily focuses on directly learning the global temporal context in the input sequence using stochastic processes with data-driven `priors'.    

\noindent \textbf{AU Intensity Estimation:} In order to model the subjective-variability of the contextual factors in AU intensity estimation, several works advocate for probabilistic models \cite{walecki2017deep, rudovic2012multi, dapogny2015pairwise, rudovic2012kernel, eleftheriadis2017gaussian, eleftheriadis2014discriminative, sandbach2013markov}. Particularly, Gaussian Processes based models \cite{eleftheriadis2017gaussian, eleftheriadis2014discriminative} and variational latent variable models~\cite{eleftheriadis2016variational, linh2017deepcoder, walecki2015variable} have been extensively used for this purpose. Other probabilistic approaches to facial affect modelling include Conditional Ordinal Random Fields (CORF)~\cite{walecki2017deep, rudovic2012multi, rudovic2012kernel} and Markov Random Fields (MRF)~\cite{sandbach2013markov}.
To learn the task-specific temporal context, context-aware feature fusion and label fusion modules are proposed in \cite{zhang2019context} to model AU intensity temporal dynamics. Although we also propose to learn the task-specific temporal context, our context learning method is completely different to that of~\cite{zhang2019context}.

\noindent \textbf{Stochastic Process Modelling}~\cite{williams2006gaussian,wang2019exact,trapp2020deep,tresp2001mixtures}: Our method primarily builds on the recently proposed stochastic regression frameworks of Neural Processes (NPs). NPs~\cite{garnelo2018conditional, garnelo2018neural, kim2018attentive, gordon2019convolutional, singh2019sequential, lee2020bootstrapping} are a family of predictive stochastic processes that aim to learn distributions over functions by adapting their priors to the data using context observations. Application of NPs in Computer Vision has been limited to basic image completion and super-resolution tasks \cite{kim2018attentive} as \textit{proof-of-concept} experiments. To our knowledge, we are the first to show how NPs can be used to solve challenging real-world regression tasks like that of emotion recognition.

\section{Method}

\def\Im{\mathcal{I}}
\def\Loss{\mathcal{L}}
\def\x{{\bf x}}
\def\y{{\bf y}}
\def\r{{\bf r}}
\def\s{{\bf s}}
\def\z{{\bf z}}
\def\W{{\bf W}}
\def\Re{\mathbb{R}}
\def\prob{\mathbb{P}}
\def\hIm{\hat{\mathcal{I}}}
\def\hI{\hat{I}}
\def\eE{\mathbb{E}}
\def\cC{\mathcal{C}}
\def\tT{\mathcal{T}}
\def\pP{\mathcal{P}}
\newcommand{\bb}[1]{\bf{#1}}
\newcommand{\m}[1]{\ensuremath{\mathcal{#1}}}
\newcommand{\h}[1]{\tilde{#1}}
\newcommand{\bbold}[1]{{[\bf #1]}}
\newcommand{\obold}[1]{{\bf #1}}

\subsection{Problem Formulation and Background}

\noindent \textbf{Temporal Regression:} Given as input signal the sequence ${\bf x}_t$ representing, for example, a sequence of images, and the corresponding output labels ${\bf y}_t \in \mathbb{R}^{\mathcal{N}}$, we wish to learn a  function from $\mathcal{X}$ to label space $\mathcal{Y}$ as $f: \mathcal{X} \rightarrow \mathcal{Y}$. 

\noindent \textbf{Stochastic Process (SP):} Unlike previous work in deterministic regression, we are not tasked with estimating a single function $f$ but rather a distribution of functions $\pP$, i.e. a \textit{Stochastic Process (SP)}, from which we can sample $f \sim \pP$. Each observed sequence in a training set is considered a realisation of the SP.

\noindent \textbf{Neural Process (NP):} NP uses an encoder-decoder architecture to learn how to sample functions $f \sim \pP$ (i.e. from an SP) \textit{conditioned} on a set of \textit{Context Points}. Specifically, a given sequence is split into two parts: the context $\cC$ for which \textit{it is assumed that the ground truth labels are available} $\{{\bf x}_c, {\bf y}_c\}_{c \in \cC}$ and the target $\tT$ points, for which only the input $\{{\bf x}\}_{t \in \tT}$ is available. 
Given a function $f$ drawn from an SP conditioned on the context $\cC$, the goal is to estimate $f(\x_t)$ for each $\x_t \in \tT$. Later on, we propose a way to deal with the unavailability of context ground truth labels $\{{\bf y}_c\}_{c \in \cC}$ for the practical scenario of emotion recognition.

\textit{The NP Encoder} is an MLP $q_\phi$ that computes an individual feature representation $\r_c = q_\phi(\x_c, \y_c) \in \mathbb{R}^d$ ($d=128$ in our work) for each input context pair, followed by a permutation-invariant aggregation operation. We choose simple summation for aggregation to compute a global deterministic context representation $\r_C = \frac{1}{C}\sum_c \r_c \in \mathbb{R}^d$, with $C$ the number of context points. To account for the context stochasticity, $\r_C$ is used to parameterise a Gaussian distribution from which a global latent variable $\z$ can be drawn. To this end, two linear layers are used to compute the mean $\mu_C(\r_C)$ and standard deviation $\sigma_C(\r_C)$. The global latent variable $\z \sim \mathcal{N}( \mu_C, \sigma_C) \in \mathbb{R}^d$ represents a different realisation of the SP conditioned on the context.

\textit{The NP Decoder} is an MLP $f_\theta$ that samples functions $f \sim P$ through the latent variable $\z$. For a given $\z$, function samples are generated at all target points as ${\bf y}_t = f_\theta(\x_t,\z)$. Assuming some i.i.d. Gaussian observation noise with learnable variance~\cite{kendall2017uncertainty}, the output of $f_\theta$ can be that of a mean and standard deviation of a Gaussian distribution: ${\bf y}_t \sim \mathcal{N}(f_\theta^\mu(\x_t, \z), f_\theta^\sigma(\x_t, \z))$.

\subsection{Affective Processes}
\label{ssec:context_selection}
\noindent Let us assume that we are given a sequence of images $\{I_t\}_{t=1}^T$, and that we want to estimate facial affect, in terms of Valence and Arousal, on a per-frame basis ${\bf y}_t \in \mathbb{R}^{2}$ \footnote{For our experiments, we will also consider the task of Action Unit intensity estimation.}. The framework of NPs cannot be applied to emotion recognition as is. Actually, application of NPs in Computer Vision has been limited to basic image completion and super-resolution tasks~\cite{kim2018attentive} as \textit{proof-of-concept} experiments. One of our main contributions is to show how such a framework can be applied for challenging real-world temporal regression tasks like the one of emotion recognition.

\subsubsection{Input Features and Pseudo-labels}

The original NP formulation would directly operate on the sequence of images $\{I_t\}$. Furthermore, it would require ground truth labels for the context points ${\bf y}_{c}$ both at training and test time. While one can assume the ground-truth labels to be available for training, their need at test time would limit the application of NPs to real-world problems. To alleviate these issues, we firstly propose to train a backbone CNN, comprising a feature extractor $\Phi(.)$ and a classifier $\W$, in a standard supervised manner, to provide \textit{independent} per-frame estimates of Valence and Arousal $\hat{\y}_t \in \mathbb{R}^{2}$. 

The purpose of this network is two-fold: Firstly, the feature extractor $\Phi(.)$ is used to provide a per-frame feature representation ${\bf x}_t = \Phi(I_t)$. Then, we can use the sequence of features ${\bf x}_t$ as input to the NP instead of the image sequence. Secondly, we propose to use the output of the classifier $\hat{\bf y}_t$ as \textit{pseudo-ground truth} labels to be fed as input to the NP. These labels are noisy, and lack temporal consistency, yet they still provide some information regarding the true labels. In this paper, we choose as our backbone an architecture inspired by that of~\cite{ntinou2021,toisoul2021estimation,yang2020fan}. We refer the reader to the Supplementary Material for further details.

\noindent \textbf{Frozen Backbone:} It is important to remark that $\Phi(.)$ and $\W$ are kept frozen during the training of our AP. While one could opt for finetuning the backbone while training the AP, the change in $\Phi(.)$ and $\W$ would lead to changing both the input space and the pseudolabels, leading to poorer performance. When minimising the AP objective (see below), $\W$ is no longer learned through the error of the values $\hat{\y}$, and thus it can lead to produce pseudolabels that lie out of the space $\mathcal{Y}$, i.e. that are not valid. Recall that NPs rely on estimating $f(\x_t)$ for the target points, provided some observed values $\y_c = f(\x_c)$ for the context. In this paper the values $\y_c$ are replaced by the predicted values $\hat{\y}_c$ provided by $\W$. If these values change over the course of training, and become non plausible values of $f$, the whole NP framework becomes invalid. We experimentally tried fine-tuning the backbone, and observed a poor performance. 

Overall, to our knowledge, we are the first to show in NP literature that pseudo-ground truth generated from an independent network can be effectively used for defining the labels for the context points used as input to the NP.

\subsubsection{Context Selection} 

Our first and obvious option for context selection is uniform random sampling from the sequence, which would imply the strong assumption that the errors of the backbone are consistent. Considering that the backbone works in a per-frame basis, it is likely that some of the input features are more indicative of the context than others. In this paper, we propose to delve into the properties of the NPs to investigate whether we can infer which frames are most optimal for context representation.

Given that the number of context points can be any number between $1$ and the length of the sequence, the choice of $\r_C = \r_c$ for a single context point $c$ also results in a valid latent function. Building on this concept and the fact that $\sigma_C(\r_C)$ represents the uncertainty of the latent functions, we propose to select the context by first estimating the set of latent functions that each of the individual representations $\r_c$ would propose, and select those with lowest uncertainty. That is, we propose to first estimate $\r_c$ and their corresponding $\sigma_C(\r_c)$ for all the frames in a given sequence, and select the $C$ with lowest uncertainty. The context set $\cC$ is then defined as the $C$ frames for which $\sigma_c(\r_c)$ is the lowest. Note that this step is used to select the context points, only. Once context points are selected, they are passed through the aggregation function to compute $\r_C$ in the standard way. This adds a negligible extra complexity to the framework (the values of $\r_c$ do not need to be computed again). Finally, we emphasize that we apply context selection only at test time.

\begin{figure}[t!]
    \centering
    \includegraphics[width=0.48\textwidth]{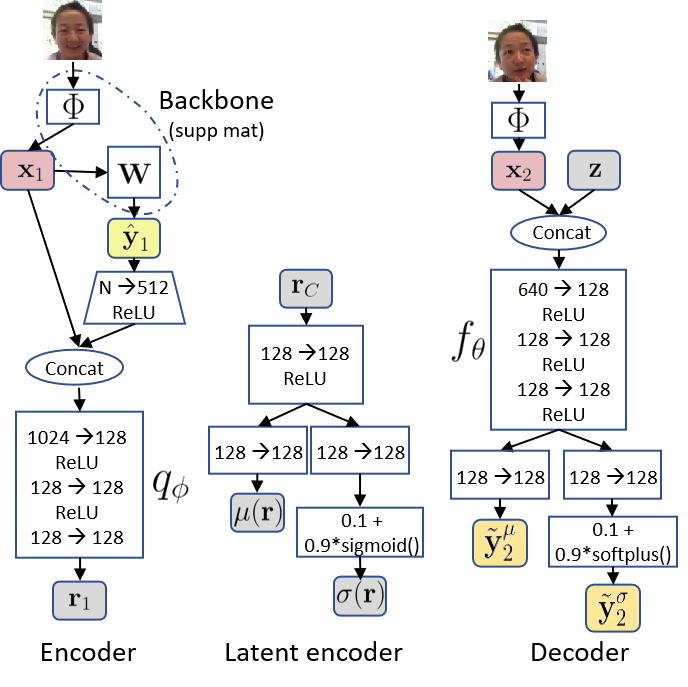}
    \caption{Architecture of AP. The encoder takes as input individual pairs $(\x_c, \hat{\y}_c)$, to produce the corresponding feature representations $\r_c$. The representations $\r_c$ for the context are summarised into $\r_C$, and passed through a latent encoder that outputs the latent distribution, defined as a Gaussian parameterised by $\mu(\r)$ and $\sigma(\r)$. The decoder takes a feature representation $\x_t$ from the target set, as well as a sample from the latent distribution, and produces an output Gaussian distribution parameterised by the outputs $(\tilde{\y}^\mu_t$, $\tilde{\y}^\sigma_t)$ of the decoder. The numbers represent the dimensions of the linear layers.}
    \label{fig:network}
\end{figure}

\subsubsection{The Complete Architecture} 

The complete architecture is shown in Fig.~\ref{fig:network}. We use as input the features $\x_t$ extracted from the pre-trained backbone, as well as the corresponding predictions. The encoder $q_\phi$ is a three-layer MLP that maps each of the input pairs $(\x_c, \hat{\y}_c)$ into a latent representation $\r_c \in \mathbb{R}^{128}$. The input pseudolabels $\hat{\y}_t$ are firstly upsampled to match the dimension of the feature representation. Once $\r_C$ has been computed from the average of the selected context points, it is forwarded to a \textit{latent encoder}, which comprises a common linear layer and two independent linear layers, to compute $\mu_C$ and $\sigma_C$, respectively. The decoder $f_\theta$ is another three-layer MLP that receives the individual features $\x_t$ and the latent representation $\z$, and produces two outputs $\tilde{\y}^\mu_t$ and $\tilde{\y}^\sigma_t$, parameterising the output distribution. During training, $\z \sim \mathcal{N}(\mu_C, \sigma_C)$, and $\tilde{\y}_t \sim \mathcal{N}(\tilde{\y}^\mu_t, \tilde{\y}^\sigma_t)$. At test time, we use the predictive mean and set $\z = \mu_C$ and $\tilde{\y}_t = \tilde{\y}^\mu_t$.

\subsubsection{Training Losses} 

Learning in our model comprises learning the parameters of the encoder $q_\phi$ and the decoder $f_\theta$. For a given set of annotated sequences of varying length, a random subset of frames are used as context, and the learning is the task of reconstructing the labels for the target points. In the NPs, the goal is to maximise the evidence lower-bound (ELBO) of $\log p(\y_T | \x_T, \x_C, \y_C)$. 
Because Valence and Arousal take values in $\left[-1,1\right]$, we also consider that the labels $\y$ are normally distributed around the origin, and propose a regularisation term for the output distribution as $\mathcal{L}_{reg} = D_{KL}( \mathcal{N}( \sum\tilde{\y}^\mu_t, \sum\tilde{\y}^\sigma_t) \, \| \, \mathcal{N}(0,1))$. 

The full objective to minimise is:
\begin{equation}
    \mathcal{L} = \mathcal{L}_{nll} + \lambda_{kl} \mathcal{L}_{kl} +  \lambda_{reg} \mathcal{L}_{reg},
    \label{elbo}
\end{equation}
with $\mathcal{L}_{nll} = -\mathbb{E}_{q_\phi(\z|\cC)}[\log p(\y_T | \x_T,\z)]$ and $\mathcal{L}_{kl} = D_{KL}(q_\phi(\z|\tT) \| q_\phi(\z | \cC))$. The optimisation can be efficiently done through the reparametrization trick~\cite{kingma2013auto}. In this paper $\lambda_{kl}$ and $\lambda_{reg}$ are set to $1$.

\subsection{Comparison with other methods}
\label{ssec:comparison}
\noindent It is important to remark that our model comes with the following benefits: a) it allows to automatically infer the global context; this adds a new key to existing Neural Processes where context is manually given, b) it allows to model the uncertainty on the estimated context, and c) it allows to draw coherent samples; note that while the decoder works on a per-frame basis, the fact that it is conditioned on a global latent variable whose value can be drawn from a latent distribution, makes all of the samples be dependent on it~\cite{de1937prevision}. 

\noindent \textbf{Vs. RNN:} The task of temporal regression could be approached by using a parametric form for $f$, by means of an RNN. However, this assumes that the variables ${\bf y}_t$ are drawn from Markov chains, which in practice disregard long-term dependencies needed for temporal context modelling. Furthermore, it can be shown~\cite{garnelo2018neural} that the joint distribution $p(\y_T)$ is exchangeable, i.e. it is permutation invariant. This offers a new insight into temporal modelling that differentiates from RNNs that condition a current observation on the previous hidden state, i.e. where order matters. We argue that, provided a given static image, the corresponding labels should not be position-dependent. If we shuffle a given sequence and pick up a frame at random, the corresponding annotation is not expected to change. Our approach seeks an orderless modelling of sequences, by looking at them as a whole through the global latent variable $\z$. 

\noindent \textbf{Vs. Self-attention:} Self-attention models~\cite{vaswani2017attention} can also model long-term dependencies using pair-wise feature similarities. On the contrary, we choose NPs to estimate $f$ by sampling from a global latent random variable. This allows for a more direct and holistic modelling of the temporal context. Furthermore, unlike self-attention models, in NPs both input signal $\x_c$, and labels $\y_c$ are used to learn context.

\section{Experimental Settings}
\label{sec:experiments}

\subsection{Datasets and Performance Metrics}
\noindent We validate our approach for emotion recognition on \textbf{SEWA}~\cite{kossaifi2019sewa} and \textbf{Aff-Wild2}~\cite{kollias2017recognition,kollias2018aff,kollias2018multi,kollias2019deep,kollias2019expression,kollias2020analysing,zafeiriou2017aff}. We also validate our approach for the task of Action Unit intensity estimation on \textbf{BP4D}~\cite{zhang2014bp4d, valstar2015fera} and \textbf{DISFA}~\cite{mavadati2013disfa}.

\noindent \textbf{SEWA}~\cite{kossaifi2019sewa} is a large video dataset for affect estimation in-the-wild. It comprises over $2,000$ minutes of video data annotated with valence and arousal values. It contains subjects from $6$ different cultures. We follow the evaluation protocol of ~\cite{kossaifi2019sewa,Kossaifi_2020_CVPR}, and apply a $8$:$1$:$1$ dataset partition\footnote{The same partitions were kindly provided by the data owners.}. 

\noindent \textbf{AffWild2}~\cite{kollias2020analysing} is a large-scale video dataset divided in training, validation, and test partitions. Because the labels for the latter are not made publicly available, we use the training and validation only for our evaluation purposes. 

\noindent \textbf{BP4D}~\cite{zhang2014bp4d, valstar2015fera} contains videos collected from 41 subjects, performing 8 different tasks. It is the main corpus of FERA 2015 challenge \cite{valstar2015fera}. We use the official train ($168$ videos), validation ($160$ videos), and test ($159$ videos) partitions, which contain annotations for 5 different AUs. 

\noindent \textbf{DISFA} \cite{mavadati2013disfa} contains 27 videos of 27 different subjects, with 4,844 frames each. It contains annotations for 12 AUs. We follow a 3-fold cross-validation protocol with 18 subjects used for training and 9 subjects used for testing. 

%\subsection{Performance metrics} 

\noindent \textbf{Performance Metrics:} For Valence and Arousal, we follow the AVEC Challenge series~\cite{ringeval2019avec} and report the Concordance Correlation Coefficient~\cite{lawrence1989concordance}. For Action Units, we follow the ranking criteria used in FERA challenges~\cite{valstar2015fera}, and report the Intra Class Correlation (ICC~\cite{shrout79}). A full definition of these metrics can be found in the Supp. Material.

\subsection{Implementation details} 
\noindent All experiments are developed in PyTorch~\cite{paszke2017}. First, we pre-train the backbone architecture $\Phi$ and $\W$ on a per-frame basis for each corresponding dataset (see Supplementary Material). We then use as input the features $\x_t$ extracted from the pre-trained backbone, as well as the corresponding predictions. The dimension of input features $\x_t$ is $512$. 

\noindent \textbf{Valence and Arousal:} To train the AP for Valence and Arousal, we randomly sample sequences between $35$ and $70$ frames from the corresponding training set. The range $35-70$ can be seen as an augmentation parameter: using a shorter range yielded worse results due to lack of generalisation, whereas no improvement was observed when using a longer range of sequences. The number of context points is randomly chosen from $3$ to the length of the given sequence. During training, the target labels are the ground-truth annotations. For the context labels, we use the ground-truth annotations or the predictions from $\W$ with probability $0.5$. The batch size is $16$ videos, and the learning rate and weight decay are set to $0.00025$ and $0.0001$, respectively. The AP is trained for $25$ epochs, each of $1000$ iterations, using Adam~\cite{kingma2014adam} with $(\beta_1,\beta_2) = (0.9, 0.999)$. We use a Cosine Annealing strategy for the learning rate~\cite{loshchilov2016sgdr}. At test time, we use a fixed sequence length of $70$ frames.

\noindent \textbf{Action Units:} We used the same setting as for Valence and Arousal, with the following differences: the batch size is set to $6$ videos, and the learning rate and weight decay are set to $0.0001$ and $0.0005$, respectively. 

\section{Ablation Studies}
\label{ssec:ablation}
\noindent
We use the validation partition of SEWA to perform our ablation studies. We study the effect of our proposed context selection method, the contribution of the pseudo labels, the effect of the different losses used for the AP training, and the effect of the latent variable. Finally, we explore a different alternative to the AP configuration~\cite{le2018empirical}. We study in Sec.~\ref{ssec:sota} the complexity of our approach.

\begin{figure}[t!]
    \centering
    \includegraphics[width=0.45\textwidth]{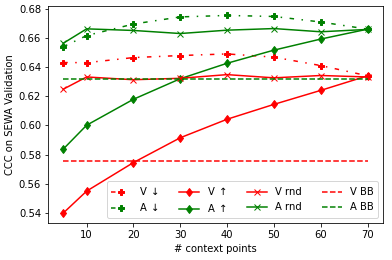}
    \caption{Accuracy on SEWA validation w.r.t. the number of context points and context selection method. The results in red correspond to the accuracy obtained for Valence (V), and those in green for Arousal (A). $\downarrow$ (dashed and dotted lines) represents the results obtained by selecting the context frames according to the lowest values of $\|\sigma_c\|$, and $\uparrow$ (solid lines with diamond markers) represents the results obtained by selecting the frames according to the largest values of $\|\sigma_c\|$ and, $rnd$ (solid lines with cross markers) represents the results obtained by choosing the frames randomly. The dashed horizontal lines represent the results of the backbone. }%\vspace{-12pt}
    \label{fig:sewa_context}
\end{figure}

\paragraph{Effect of context selection:} Firstly, we investigate the effect of the number of selected context points and the selection method proposed in Sec.~\ref{ssec:context_selection}. For a given trained model, we study the effect of choosing the context points (a) randomly, (b) according to the lowest values of $\|\sigma_c\|$, and (c) according to the largest values of $\|\sigma_c\|$. Fig.~\ref{fig:sewa_context} shows that the best results are attained when choosing 40 context points according to the lowest value of $\|\sigma_c\|$. We fix for the remaining experiments the number of context points to $40$.
%\vspace{-2pt}
\paragraph{Impact of task-aware context modelling:} One of the contributions of our method is that of task-aware context modelling in the form of \textit{pseudo-ground truth labels}. We conducted an experiment where the AP model is trained without the predictions provided by $\W$. The architecture, training, losses, learning rate and weight decay remain the same as for the main AP. The results, reported in Table~\ref{Tab:sewa_ablation} as $\text{AP}_{\text{w/o } \W}$, show that the performance drops significantly without the aid of the noisy task-specific predictions.  

\begin{table}[h!]
\begin{center}
\begin{tabular}{c c c c}
\hline
\textbf{Model} & \textbf{Valence} & \textbf{Arousal} & \textbf{Mean} \\
\hline
Backbone & 0.576 & 0.632 & 0.604 \\
$\text{AP}_{nll}$ & 0.645 & \textbf{0.676} & 0.660 \\
$\text{AP}_{nll+kl}$ & 0.649 & 0.669 & 0.659 \\
$\text{AP}_{nll+reg}$ & 0.649 & \textbf{0.676} & 0.662 \\
$\text{AP}_{nll+reg+kl}$ & 0.643 & 0.628 & 0.635 \\
$\text{Deterministic AP}$ & 0.650 & 0.576 & 0.613 \\
$\text{AP}+\text{Det.}$ & 0.649 & 0.658 & 0.652 \\
$\text{AP}_{\text{w/o }\W}$ & 0.600 & 0.454 & 0.527 \\
$\text{AP}+\text{Det.} + \text{Att}$ & \textbf{0.662} & 0.672 & \textbf{0.667} \\
\hline
\hline
\end{tabular}
\end{center}
\caption{Ablation studies on the validation set of SEWA.} %\vspace{-13pt} 
\label{Tab:sewa_ablation}
\end{table}
%\vspace{-2pt} 

\paragraph{Effect of global latent random variable:} We firstly investigate the effect of the latent variable on regressing coherent functions. Recall that our proposed approach consists of a probabilistic regression of coherent functions, i.e. it produces a set of functional proposals. An alternative to this approach is to use a non-probabilistic approach where the latent representation does not parameterise a latent random distribution, but is rather \textit{deterministic}. The architecture of a deterministic AP is similar to that shown in Fig.~\ref{fig:network}, although with a single-branch latent encoder and no sampling. With a deterministic AP, one dispenses with the coherent sampling i.e. the dependence among the target outputs is lost, and assumed to be i.i.d. The learning burden is put only into capturing the observation variance $\tilde{\y}^\sigma$ to account for the output uncertainty (i.e. it is probabilistic in the output, but not in the latent representation). The results of the deterministic AP are shown in Table~\ref{Tab:sewa_ablation} (Deterministic AP). We observe that the performance degrades substantially, in particular regarding the Arousal predictions. This illustrates the importance of the latent variable to model the underlying variability for the joint prediction of Valence and Arousal. Finally, we also investigated the dimensionality of $\z$, observing little effect when having more dimensions. 

\paragraph{Effect of losses:} The training of the APs is carried out by minimising the objective shown in Eqn.~\ref{elbo}. Recent findings~\cite{foong2020meta} suggest that training an NP using only a maximum likelihood approach (i.e. by minimising the negative log-likelihood $\mathcal{L}_{nll}$) can improve performance and avoid the practical issues of using amortised variational inference (i.e. $\mathcal{L}_{nll} + \mathcal{L}_{kl}$). We studied the effect of each case along with our proposed loss. The results are shown in Table~\ref{Tab:sewa_ablation}. We refer with $\text{AP}_{nll}$ to the maximum likelihood approach, with $\text{AP}_{nll+kl}$ to the one that minimises $\mathcal{L}_{nll} + \mathcal{L}_{kl}$, with $\text{AP}_{nll+reg}$ to the one that minimises $\mathcal{L}_{nll} + \mathcal{L}_{reg}$, and with $\text{AP}_{nll+reg+kl}$ to the one that minimises the full loss in Eqn.~\ref{elbo}. The numerical results do not shed light into whether one is better than the other. We observe that a combination of the maximum likelihood approach with our proposed regularisation slightly improves the numerical results, although we acknowledge these not to be significant enough. If we use the three losses together we observe a performance degradation. These results align with recent findings that advocate for a further investigation of which of the losses is a better objective for Neural Processes~\cite{foong2020meta}.

\begin{figure*}[t!]
    \centering
    \includegraphics[width=0.99\textwidth]{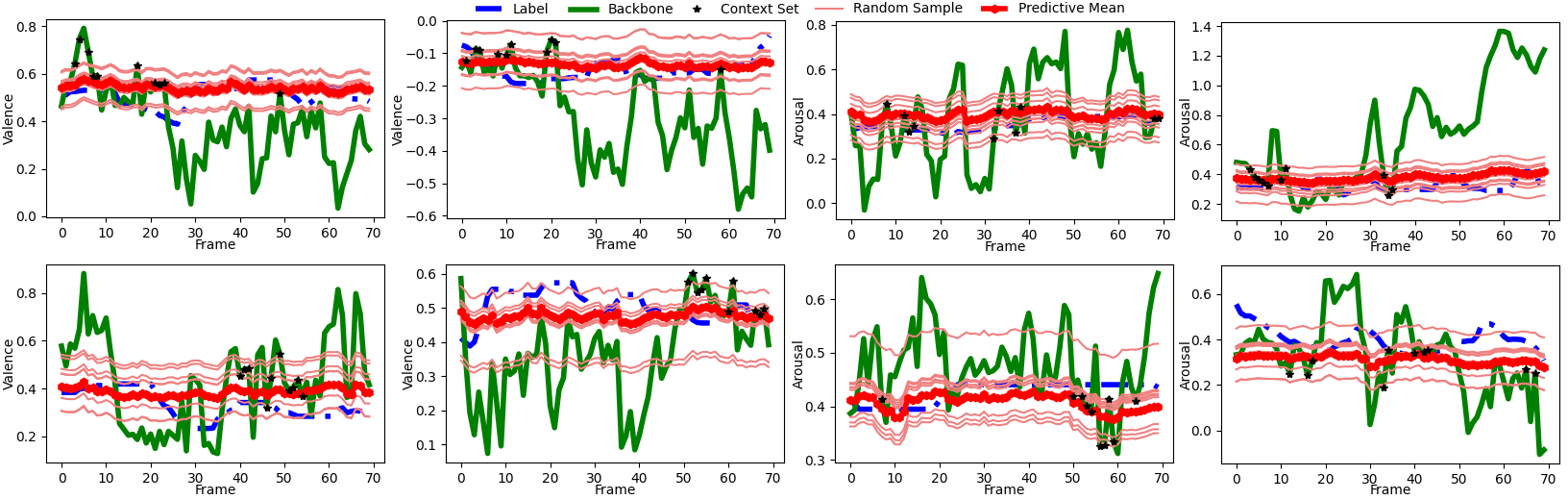}
    \caption{Examples of modelling Valence and Arousal on SEWA dataset, when using as context only $10$ points, selected according to the lowest values of $\|\sigma_c\|$. Blue line represents the ground-truth labels. The green line represents the predictions from the backbone. The black dots represent the frames selected as context using $\sigma_c$. The light red curves represent different realisations of $f(.,\z)$ for $10$ different (sampled) $\z$, and the red curve represents the process returned by the predictive mean $\mu_C$. The variation in the functions corresponds to the global latent uncertainty given the selected context. }
    \label{fig:sampling}
\end{figure*}

\paragraph{Different network configurations:} 

We also study different network configurations. 
We first evaluate the performance of using a combined latent and deterministic layer denoted as $\text{AP}+\text{Det.}$. The performance of the $\text{AP}+\text{Det.}$ is on par with that of the single AP. We then explore the use of Attention in the $\text{AP}+\text{Det.}$ where self-attention is used to replace the aggregation of the individual representations (attention is permutation-invariant). 
Considering the trade-off between complexity and accuracy, we opt for keeping the simplest architecture, described in Fig.~\ref{fig:network}.

\begin{table*}[t!]
    \begin{minipage}{.5\linewidth}
      \begin{center}

        \begin{tabular}{c c c c}
\hline
\textbf{Model} & \textbf{Valence} & \textbf{Arousal} & \textbf{Mean} \\
\hline
Mitekova et al.~\cite{mitenkova2019valence} & 0.44 & 0.39 & 0.42 \\
Kossafi et al.~\cite{Kossaifi_2020_CVPR} &\textbf{0.75} & 0.52 & 0.64 \\
Ours backbone & 0.69 & 0.61 & 0.65 \\
ConvGRU \cite{cho2014learning}$^\dagger$ & 0.72 & 0.62 & 0.67 \\
Self-Attn \cite{vaswani2017attention}$^\dagger$ & 0.70 & 0.61 & 0.65 \\
Ours AP & \textbf{0.75} & \textbf{0.64} &\textbf{0.69} \\
\hline 
\hline
\end{tabular}
 \vspace{5pt}
\caption{ Comparison against state of the art methods on the \textbf{SEWA} test partition. $^\dagger$ denotes in-house implementation.} 
\label{Tab:sewa_benchmarks}
\end{center}
    %\end{subtable}%
    \end{minipage}% 
    \hfill
    \begin{minipage}{.47\linewidth}
      \centering
\begin{tabular}{c c c c}
\hline
\textbf{Model} & \textbf{Valence} & \textbf{Arousal} & \textbf{Mean} \\
\hline

Ours Backbone & 0.398 & 0.445 & 0.422 \\
ConvGRU~\cite{cho2014learning}$^\dagger$ & 0.398 & 0.503 & 0.450  \\
Self-Attn~\cite{vaswani2017attention}$^\dagger$ & 0.419 & \textbf{0.505} & 0.462 \\
Ours AP & \textbf{0.438} & 0.498 & \textbf{0.468} \\
\hline
\hline
\end{tabular} 
 \vspace{5pt}
\caption{Comparison of different methods on the validation set of \textbf{AffWild2}. $^\dagger$ denotes in-house implementation.} \label{Tab:affwild2_benchmarks}
    \end{minipage}  \\

\begin{minipage}{1\linewidth}
    \vspace{8pt}
      \centering
        \begin{tabular}{c c c c c c c c c c c c c c}
\hline
\textbf{AU} & \textbf{1} &  \textbf{2} &  \textbf{4} &  \textbf{5} &  \textbf{6} &  \textbf{9} &  \textbf{12} &  \textbf{15} &  \textbf{17} & \textbf{20} &  \textbf{25} &  \textbf{26} & \textbf{Avg.} \\
\hline
VGP-AE ~\cite{eleftheriadis2016variational}  &  0.48    &  0.47    &  0.62  & 0.19    & 0.50   & 0.42 & 0.80 & 0.19 & 0.36 & 0.15 & 0.84 & 0.53 & 0.46 \\
2DC  ~\cite{linh2017deepcoder}   & \textbf{0.70}  &  \textbf{0.55} &  0.69  &  0.05   &\textbf{0.59}  & 0.57 &\textbf{0.88}& 0.32 & 0.10 & 0.08 & 0.90 & 0.50 & 0.50\\
HR ~\cite{ntinou2021}    &  0.56    &  0.52    &  0.75  &  0.42   &  0.51  & 0.55 & 0.82 &\textbf{0.55}& 0.37 & 0.21 &\textbf{0.93}& 0.62 & 0.57 \\
Ours backbone               & 0.25     &  0.15   &  0.76  &    0.77	 & 0.50  & 0.62 & 0.82 & 0.44 & 0.58 & 0.22 &\textbf{0.93}&\textbf{0.69}& 0.56                		 \\
BiGRU    ~\cite{cho2014learning}$^\dagger$                           & 0.23     &  0.13    &  0.77 &	0.70   & 0.53  & 0.64 &  0.82 & 0.42 & 0.58 & 0.25 & 0.92 &\textbf{0.69}&  0.56              			\\
Self-Attn ~\cite{vaswani2017attention}$^\dagger$                         & 0.30     &  0.17   & \textbf{0.78}& \textbf{0.76}  & 0.51  & 0.64 & 0.82 & 0.46 & 0.59 & 0.17 & 0.92 & 0.68 & 0.57               				\\
Ours AP    & 0.35     &  0.19   & \textbf{0.78}&    0.73  & 0.52 & \textbf{0.65}& 0.81 & 0.49 &\textbf{0.61}&\textbf{0.28}& 0.92 & 0.67 & \textbf{0.58}              		\\
\hline
\hline
\end{tabular} 
 \vspace{2pt}
\caption{Comparison against state of the art methods on the \textbf{DISFA} database (in ICC values). $^\dagger$ denotes in-house evaluation} \label{Tab:disfa_benchmarks}
    \end{minipage}
\end{table*}

\begin{table}[h!]

\begin{center}
\begin{tabular}{c c c c c c c}
\hline
\textbf{AU} & \textbf{6} & \textbf{10} & \textbf{12} & \textbf{14} & \textbf{17} & \textbf{Avg.} \\
\hline
CDL \cite{baltruvsaitis2015cross}             & 0.69 & 0.73 & 0.83 & 0.50 & 0.37 & 0.62 \\ 
ISIR \cite{nicolle2015facial}                       & 0.79 & 0.80 & \textbf{0.86} & \textbf{0.71}  & 0.44 & 0.72 \\
HR~\cite{ntinou2021}                                & \textbf{0.82} & \textbf{0.82} & 0.80 & \textbf{0.71} & 0.50 & 0.73 \\
Ours backbone        & 0.76 & 0.74 & 0.82 &  0.62 &  0.39 & 0.67 \\
BiGRU~\cite{cho2014learning}$^\dagger$               & 0.78 & 0.76 & 0.83 & 0.62 & 0.50 &  0.70 \\
Self-Attn~\cite{vaswani2017attention}$^\dagger$  & 0.80 & 0.78 & 0.79 & 0.60 & 0.42 & 0.68 \\
Ours AP	& \textbf{0.82} & 0.80 & \textbf{0.86} & 0.69 & \textbf{0.51} &\textbf{0.74}\\

\hline
\hline
\end{tabular}
\end{center}
\caption{Results on the test partition of \textbf{BP4D} dataset (in ICC values). $^\dagger$ denotes in-house evaluation} \label{Tab:bp4d_benchmarks}%\vspace{-10pt}
\end{table}

%\label{ssec:qualitative}
\noindent \textbf{Qualitative results:}  We show a conceptual demonstration of the AP capabilities for both context selection and functional proposals in Fig.~\ref{fig:sampling}. The blue lines represent the ground-truth annotations for different segments of $70$ frames. The green lines correspond to the individual predictions $\tilde{\y}_t$ given by the backbone $\W$. The black dots represent the frames selected as context according to the lowest values of $\|\sigma_c\|$. For the sake of clarity, we used only $10$ points to select context. The light red curves represent different realisations of the functions $f_\theta(.,\z)$ given by different samples of $\z$, with the function corresponding to $\mu_C$ represented in dark red. We can see that the proposed functions are able to better approximate the sequential predictions from the noisy individual representations, leading to much more stable predictions. We can observe that the model tends to select frames that correspond to lower error, indicating how the individual latent representations can represent their confidence in the context.

\section{Comparison with state-of-the-art}
\label{ssec:sota}
\noindent We compare the results of our models w.r.t. the most recent works in emotion and Action Unit intensity estimation. The bulk of our results are shown in Tables~\ref{Tab:sewa_benchmarks} and~\ref{Tab:affwild2_benchmarks} for Valence and Arousal, and Tables~\ref{Tab:disfa_benchmarks} and~\ref{Tab:bp4d_benchmarks} for Action Units (MSE is also reported in the Supplementary Material). 

\paragraph{In-house baselines:} We trained a set of baselines based on GRU~\cite{cho2014learning} and Attention~\cite{vaswani2017attention}, using, for each case, the same backbone as the one used in AP to provide the input features. We also optimised for each task and model the sequence length (herein denoted by $L$). For Valence and Arousal, we used a $4$-layer Convolutional GRU with $128$-d hidden layers, and $L=30$. For Action Units, we trained a $2$-layer Bidirectional GRU with $256$-d hidden states, with $L=60$ frames. For Attention, we used, for both tasks, an architecture based on the Transformer encoder: we used a $2$-layer MultiHead Self-Attention encoder with $32$ learnable heads, each of $512$ dimensions. The sequence lengths are $L=30$ and $L=40$ frames for Valence and Arousal and Action Units, respectively. The results in Tables~\ref{Tab:sewa_benchmarks}-~\ref{Tab:bp4d_benchmarks} indicate that our AP outperforms both recurrent and attention-based models. We also experimented with using as input both the features and the labels, as well as finetuning the backbone, observing a performance degradation.

\paragraph{Different backbones:} We validate that our AP can be used with other backbone choices. On SEWA, we first train a ResNet-50, yielding a CCC of $0.567$, and then train our AP on top yielding a CCC of $0.655$. On AffWild-2, we train our AP using as backbone the publicly available model of~\cite{kuhnke2020two}: the backbone of~\cite{kuhnke2020two} reports a CCC score on the AffWild-2 test set of $0.43$, whereas our AP yields a CCC score of $0.47$.  

\paragraph{Analysis:} On SEWA, we compare against Mitekova et al.~\cite{mitenkova2019valence} and Kossaifi et al.~\cite{Kossaifi_2020_CVPR}. Notably, our backbone is already on par with the state of the art results of \cite{Kossaifi_2020_CVPR}. We can observe that our method based on temporal context offers the largest improvement. The same behaviour is observed on AffWild2, where our AP significantly improves the Valence CCC. On DISFA, our backbone is again on par with the state of the art methods VGP-AE~\cite{eleftheriadis2016variational} and 2DC~\cite{linh2017deepcoder}. However, our AP brings the largest improvement setting a new state-of-the-art ICC of $0.58$, which improves over the Heatmap Regression method of Ntinou et al.~\cite{ntinou2021}. Finally, we compare our method against the winner and runner-up methods of FERA 2015, ISIR~\cite{nicolle2015facial} and CDL~\cite{baltruvsaitis2015cross}, respectively, as well as against the Heatmap Regression method of~\cite{ntinou2021}. Our method sets up a new state-of-the-art result with an ICC score of $0.74$. We include a set of qualitative results in the Supplementary Material.

\paragraph{Complexity:} Our AP adds very little complexity to the pre-trained backbone. The configuration shown in Fig.~\ref{fig:network} comprises only ${\sim}330K$ parameters, and requires only ${\sim}20$ MFLOPs for a given sequence of $L=70$ frames\footnote{We measured the complexity using the Thop library for PyTorch \url{https://github.com/Lyken17/pytorch-OpCounter}}. In comparison, the ConvGRU model has ${\sim}789K$ parameters, and requires ${\sim}758$ MFLOPs for $L=30$ frames. The PyTorch built-in 2-layer BiGRU has $\sim1.2M$ parameters, and needs ${\sim}71$ MFLOPs for $L=60$ frames. The Multi-Head attention model has ${\sim}4.2M$ parameters, and requires ${\sim}125$ MFLOPs for $L=30$ frames. 
\section{Conclusions}

\noindent We proposed a novel temporal context learning approach to facial affect recognition, by modelling the stochastic nature of affect labels. To model uncertainty in the temporal context of the affective signals, our proposed method, called Affective Processes, builds on Neural Processes to learn a distribution of functions. Our method addresses a key limitation of the NPs --- that of requiring ground-truth labels at inference time --- by proposing the use of pseudo-labels provided by a pre-trained backbone and by using a novel method for automatic context selection. We believe that the key elements of our method pave the way to apply NPs to large-scale supervised learning problems which are inherently stochastic and for which deterministic temporal models like RNNs and self-attention could fall short.

\subsection*{Acknowledgements}
\noindent The work by Michel Valstar was part-funded by the National Institute for Health Research. The views represented are the views of the
authors alone and do not necessarily represent the views of the Department of Health in England, NHS, or the National Institute for Health Research.
{\small
\bibliographystyle{ieee_fullname}
\bibliography{cvpr}
}
\newpage
\appendix
\section{Backbone}
\label{sec:backbone}
\noindent This appendix is used to describe the backbone structure and training, left out of the main document due to lack of space. It also includes a formal definition of the main performance metrics used in the paper, as well as the Mean Squared Error reported for the Action Unit intensity estimation task. We first describe the architecture of the backbone (Sec.~\ref{ssec:architecture}), and then the training details for each of the databases used in the paper (Sec.~\ref{ssec:training}). Sec.~\ref{ssec:metrics} and Sec.~\ref{ssec:mse} are devoted to describing the performance metrics used in the paper and to reporting the MSE results on DISFA and BP4D, respectively.

\subsection{Architecture}
\label{ssec:architecture}
\begin{figure*}[t!]
    \centering
    \includegraphics[width=0.90\textwidth]{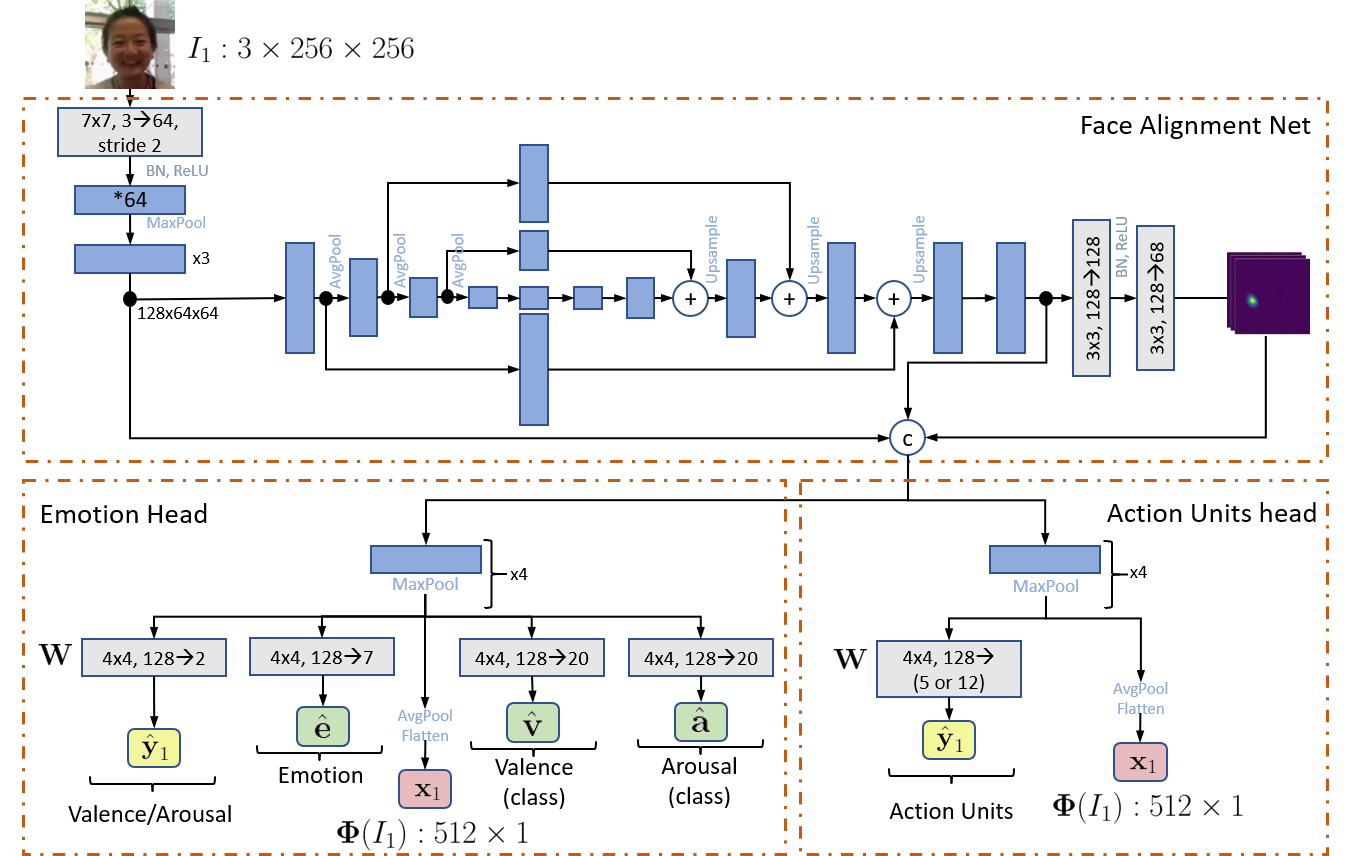}
    \caption{Architecture of the Backbone used in our AP pipeline described in the main document. For the sake of clarity, both the Emotion Head and Action Units Head are depicted together, despite these being different networks, trained separately. The grey modules represent $2$-d convolutions (Conv2d), whereas the blue blocks represent Convolutional Blocks (ConvBlocks), described in Fig.~\ref{fig:backbone}. The $^*64$ inscribed in the first ConvBlock corresponds to a slightly different configuration that uses a skip connection to upsample the number of channels from $64$ to $128$. The backbone includes a Face Alignment Network, an Hourglass-like architecture that takes the input image $I$, and produces a set $68$ Heatmaps corresponding to the position of the facial landmarks. The Hourglass comprises four layers of ConvBlocks with downsampling, and skip connections (for the sake of clarity we illustrate three layers, where each smaller block corresponds to halving the spatial resolution). As shown in Fig.~\ref{fig:convblock}, our ConvBlocks are of $128$ channels, rather than the original $256$ used in \cite{bulat2017far}. The Face Alignment Network is pre-trained and kept frozen, and returns a set of features resulting from concatenating the output of the last ConvBlock before the Hourglass, the output of the last ConvBlock of the network, and the produced Heatmaps. Then, the Emotion and Action Unit heads follow for each corresponding task. Both have a similar Feature Extraction Module, composed of $\times4$ ConvBlocks followed by Average Pooling. The output of this module is a $128\times4\times4$ tensor, which is used as input to the corresponding classifiers, as well as to compute the final feature representation ${\bf x}_t$ that will be used along with ${\hat{\bf y}}_t$ as input to our proposed AP. }
    \label{fig:backbone}
\end{figure*}

\noindent The structure of the backbone for both Valence and Arousal and Action Unit recognition is depicted in Fig.~\ref{fig:backbone}. Note that both are depicted in the same figure for the sake of clarity, although the corresponding subnetworks consisting of the Emotion Head or the Action Units Head are trained independently using the task specific datasets. Both networks share a common module, referred to as Face Alignment Module, which is pre-trained for the task of facial landmark localisation, and kept frozen for the subsequent training steps. For both Valence and Arousal and Action Unit estimation, the backbone is decomposed into three main components, namely \textit{a) Face Alignment Module}, \textit{b) Task-specific Feature Module}, and \textit{c) Task-specific Head}. 

\begin{figure}[h!]
    \centering
    \includegraphics[width=0.40\textwidth]{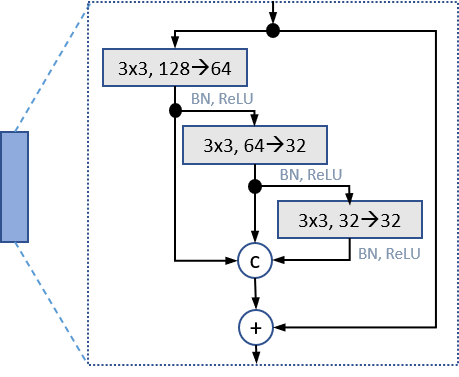}
    \caption{The Convolutional Block (ConvBlock), used in ~\cite{bulat2017far}. Instead of using $256$ channels, we opt for a lighter version and choose $128$ channels instead. }
    \label{fig:convblock}
\end{figure}

The \textit{Face Alignment Module} is a lightweight version of the Face Alignment Network of \cite{bulat2017far}. It starts with a 2d convolutional layer (referred to as Conv2d) and a set of $4$ convolutional blocks (ConvBlock, depicted in Fig.~\ref{fig:convblock}) that bring down the resolution of the input image from $256$ to $64$ and the number of channels from $3$ to $128$. This set of ConvBlocks is followed by an Hourglass, a four layer set of $128$-channel ConvBlocks with skip connections, that aggregate the features at different spatial scales. The Hourglass is followed by another ConvBlock and two Conv2d layers that produce a set of $68$ Heatmaps, corresponding to the position of the facial landmarks. In this paper, rather than using the facial landmarks to register the face, we directly concatenate the produced features at both an early and late stage of the network with the Heatmaps. The output is then a $128+128+68$ tensor of $64\times 64$, resulting from concatenating the features computed after the fourth ConvBlock, the features computed after the last ConvBlock, and the produced Heatmaps. This way, the Heatmaps help the subsequent network locally attend to the extracted coarse and fine features~\cite{ntinou2021,toisoul2021estimation,yang2020fan}. The benefits of this approach are twofold: a) it dispenses with the need of registering the faces according to detected landmarks, and b) because of a) we can directly use the features from the Face Alignment Network and have shallower networks in the front-end for the subsequent tasks.

The \textit{Task specific Feature Module} consists of a mere set of $4$ ConvBlocks, each followed by a max pooling layer, that produce a tensor of $128\times4\times4$. To form the features ${\bf x}_t$ that will be used as input to our AP network, we further downsample that tensor through an average pooling operation with a $2\times2$ kernel. The $128\times2\times2$ output is flattened to form the $512$-d feature vector ${\bf x}_t$. 

The \textit{Task specific head for Valence and Arousal} is composed of four independent Conv2d layers, each with $4\times4$ filters (i.e. equal to the spatial resolution of the input tensor). The first Conv2d layer is the corresponding Valence and Arousal classifier ${\bf W}$ mentioned in the main document. The output of this layer is a $2$-d vector ${\hat{\bf y}}_t$, corresponding to the values of Valence and Arousal, respectively. In order to boost the performance of the network for the task of predicting the continuous values of Valence and Arousal ($\hat{\bf y})$, we approach the backbone training in a Multi-task manner (see below), where the goal is to also classify the \textit{basic (discrete) emotion}, as well as the bin where both Valence and Arousal would lie in a discretised space. For the basic emotion (happiness, sadness, fear, anger, surprise, disgust and neutral), we include a second Conv2d which outputs the logits corresponding to each of the $7$ target classes. For the discretised Valence ($\hat{\bf v}$) and Arousal ($\hat{\bf a}$), we use two Conv2d layers with $20$ outputs each, i.e. we discretise the continuous space in $20$ bins, and we treat the task of predicting the corresponding bin as a classification task (see below). Note that these extra heads, as well as the emotion head, are used to reinforce the learning of the regression head tasked with predicting $\hat{\bf y}$. Once the network is trained, the heads corresponding to the discrete emotion and the discretised Valence and Arousal are removed from the backbone.

The \textit{Task specific head for Action Unit intensity estimation} is also composed of $4$ ConvBlocks as for the Valence and Arousal head. The output features, a tensor of $128\times4\times4$ are also spatially downsampled with average pooling and flattened to form the input features to the AP ${\bf x}_t$. The Action Units classifier ${\bf W}$ is a Conv2d with a $4\times4$ filter that maps the $128\times4\times4$ into either the $5$ or $12$ target AUs, for BP4D and DISFA, respectively.

\subsection{Training}
\label{ssec:training}
\paragraph{Data processing} The faces are first cropped according to a face bounding box, provided by the off-the-shelf face detector RetinaFace ~\cite{deng2019retinaface}. Given that the first block of the backbone is a Face Alignment Network that is used to provide the features to the subsequent networks, no face registration step is applied. During training, for image augmentation we applied random cropping (224$\times$224), random horizontal flipping, random rotation($-20^{\circ}$ to $+20^{\circ}$), color jittering, and random gray scaling operations. 

\paragraph{Face Alignment Network} The Face Alignment Module was trained on the 300W-LP dataset~\cite{bulat2017far} using standard Heatmap Regression, and was kept frozen afterwards. 

\paragraph{Valence and Arousal} For \textit{Valence and Arousal}, the network is trained in a Multi-task way. Let $\hat{\bf y} = (\hat{y}_v, \hat{y}_a)$ be the Valence and Arousal prediction, $\hat{\bf e} \in \mathbb{R}^7$ be the output of the discrete emotion layer, and $\hat{\bf v} \in \mathbb{R}^{20}$ and $\hat{\bf a} \in \mathbb{R}^{20}$ the output of the Valence and Arousal classes, respectively. We denote by ${\bf y}$ and ${\bf e}$ the corresponding Valence and Arousal and Emotion ground-truth values. The loss is defined as:

\begin{equation}
\begin{split}
    \mathcal{L} =& \lambda_{mse} \mathcal{L}_{mse} + \lambda_{ccc}\mathcal{L}_{ccc} \\
    & + \lambda_{xent-emo} \mathcal{L}_{xent-emo}  \\
    & + \lambda_{xent-va}\mathcal{L}_{xent-va}
\end{split}
\end{equation}
where $\mathcal{L}_{mse} = \| \hat{\bf y} - {\bf y} \|$ is the standard MSE loss for Valence and Arousal, $\mathcal{L}_{ccc} = 1 - \frac{CCC({\hat{y}}_v,{y}_v) + CCC({\hat{y}}_a,{y}_a)}{2}$ is the CCC score between the predicted Valence and Arousal values and corresponding ground-truth, $\mathcal{L}_{xent-emo}$ is the standard cross entropy loss between the predicted emotion $\hat{\bf e}$ and the corresponding ground-truth ${\bf e}$. We define $\mathcal{L}_{xent-va} = \mathcal{L}_{xent-v} + \mathcal{L}_{xent-a}$, with $\mathcal{L}_{xent-v}$ the cross entropy loss between the $20$-d output of the Valence head and the corresponding ground-truth bin, and equivalently $\mathcal{L}_{xent-a}$ for Arousal. The ground-truth bin results from uniformly discretising the Valence and Arousal spaces, which lie within the $[-1,1]$ space, into $20$ bins each. 

\begin{table*}[t!]
\begin{center}
\begin{tabular}{c c c c c c c c c c c c c c}
\hline
\textbf{AU}                                      								   &   1    &   2    &  4     &  5     &  6    &  9     &  12    &  15   &  17   &  20  &  25    &  26  & Avg. \\
\hline
VGP-AE ~\cite{eleftheriadis2016variational}      & 0.51 & \textbf{0.32} & 1.13 & 0.08 & 0.56 & 0.31 & 0.47 & 0.20 & 0.28 & \textbf{0.16} & 0.49        & 0.44        & 0.41 \\
2DC  ~\cite{linh2017deepcoder}                  		   & \textbf{0.32} & 0.39 & 0.53 & 0.26 & 0.43 & 0.30 & \textbf{0.25} & 0.27 & 0.61 & 0.18 & 0.37      & 0.55       & 0.37 \\   
HR ~\cite{ntinou2021}	                         		           & 0.41 & 0.37 & 0.70 & 0.08 & 0.44 & 0.30 & 0.29 & 0.14 & 0.26 & \textbf{0.16} & 0.24          & 0.39          & 0.32 \\   
Ours backbone                                    					&  0.93 & 0.90 & 0.51 & 0.04 & 0.44 & 0.19 & 0.30 & 0.13 & \textbf{0.21} & 0.17 & \textbf{0.23}             & 0.29            & 0.36 \\
BiGRU  ~\cite{cho2014learning}$^\dagger$         				 &  0.85 & 0.79 & 0.48 & 0.06 & 0.47 & 0.19 & 0.34 & 0.18 & 0.23 & 0.21 & 0.30          & 0.40          & 0.37 \\ 
Self-Attn~\cite{vaswani2017attention}$^\dagger$   			 &  0.76 & 0.71 & 0.52 & 0.04 & \textbf{0.42} & 0.17 & 0.35 & 0.14 & \textbf{0.21} & 0.19 & 0.28         & 0.36        & 0.34  \\
Ours  AP                                         						  &  0.68 & 0.59 &\textbf{0.40} & \textbf{0.03} & 0.49 & \textbf{0.15} & 0.26 & \textbf{0.13} & 0.22 & 0.20 & 0.35 & \textbf{0.17} & \textbf{0.30} \\
\hline
\hline
\end{tabular}
\end{center}
\caption{Results on the DISFA database  (in MSE values)  $^\dagger$ denotes in-house evaluation }
\label{Tab:disfa_benchmarks_mae}

\end{table*}

The values of the loss weights are all set to $1$ except for the MSE loss that is set to $\lambda_{mse}=0.5$. For both SEWA and AffWild2, the training is performed for $20$ epochs, using Adam with learning rate $0.0001$, $(\beta_1, \beta_2) = (0.9,0.999)$ and weight decay $0.000001$. The learning rate is reduced by a factor of $10$ after every $5$ epochs. 
 
For AffWild2 we used the sequences that were annotated with both discrete emotion and Valence and Arousal. Considering that SEWA has not been annotated with the basic emotions, we train our SEWA backbone by extending it with the sequences of AffWild2 containing such annotations. We backpropagate w.r.t. the emotion head using images from AffWild2, and w.r.t. the remaining heads using only images from SEWA. We apply the same $8$:$1$:$1$ partition described in the paper, and choose the backbone according to the best validation CCC score. 
 
\paragraph{Action Units} For \textit{Action Unit} intensity estimation, Mean Squared Error is used as the loss function to train the corresponding models in this work (for BP4D and DISFA). The AU intensities are normalised from -1 to 1 to align with the $\mathcal{L}_{reg}$ used in the AP framework described in the main document. Adam optimizer with a learning rate of $0.0003$, $(\beta_1, \beta_2) = (0.9, 0.999)$, and an L2 weight decay of $0.00001$ is used to train the Action Unit head. To tune the initial learning rate, cyclic learning rate scheduler with a cycle length of 2 is used. After 80 epochs, the best model is selected based on the ICC score on the validation set. 

For BP4D, the model is trained using the official train/validation/test partitions. For DISFA, the model is trained using the three-fold cross validation method described in the main document, using exactly the same generated partitions. 

\section{Performance Metrics}
\label{ssec:metrics}
\noindent For Valence and Arousal, we report the Concordance Correlation Coefficient~\cite{lawrence1989concordance}, which is used to rank participants in the AVEC Challenge series~\cite{ringeval2019avec}. It is a global measure of both correlation and proximity, and is defined as:
\begin{equation}
CCC(\y, \hat{\y}) = \frac{2\sigma_\y \sigma_{\hat{\y}}\rho_{\y\hat{\y}}}{\sigma_\y^2+ \sigma_{\hat{\y}}^2 + (\mu_\y - \mu_{\hat{\y}})^2},
    \label{eq:ccc}
\end{equation}
where $\mu$, $\sigma$, and $\rho$ refer to the mean value, (co-)variance, and Pearson Correlation Coefficient, respectively. 

For Action Unit intensity, we follow the standard ranking criteria used in FERA challenges\cite{valstar2015fera}, and we report the Intra Class Correlation (ICC~\cite{shrout79}). For an AU $j$ with ground-truth labels $\{y^j_i\}_{i=1}^N$, and predictions $\{\tilde{y}^j_i\}_{i=1}^N$, the ICC score is defined as $ICC^j = \frac{W^j-S^j}{W^j+S^j}$, with $W^j = \frac{1}{N}\sum_{i} \left((y^j_i - \hat{y}^j)^2 + (\tilde{y}^j_i - \hat{y}^j)^2 \right)$, $S^j = \sum_{i} (y^j_i - \tilde{y}^j_i)^2$, and $\hat{y}^j = \frac{1}{2N}\sum_i (y^j_i + \tilde{y}^j_i)$.

\section{Mean Squared Error Results} \label{ssec:mse}
\noindent The additional Mean Squared Error results for DISFA and BP4D are reported in Table~\ref{Tab:disfa_benchmarks_mae} and Table~\ref{Tab:bp4d_benchmarks_mae}. 

\begin{table}[ht]
\begin{center}
\begin{tabular}{c c c c c c c}
\hline
\textbf{AU} & \textbf{6} & \textbf{10} & \textbf{12} & \textbf{14} & \textbf{17} & \textbf{Avg.} \\
\hline
%CDL ~\cite{baltruvsaitis2015cross}                  &  -   &  -   &  -   &  -   &  -   &  -   \\ 
ISIR ~\cite{nicolle2015facial}                      		& 0.83 & 0.80 & 0.62 & 1.14 & 0.84 & 0.85  \\
HR ~\cite{ntinou2021}                                		 & \textbf{0.68}  & \textbf{0.80} & 0.79 & \textbf{0.98} & 0.64 & 0.78\\                           
Ours backbone                                       		   & 0.80 & 0.87 & 0.74 & 1.23 & 0.89 &  0.90 \\
BiGRU ~\cite{cho2014learning}$^\dagger$             			& 0.79 & 0.85 & 0.76 & 1.19 & 0.78 & 0.87 \\
Self-Attn ~\cite{vaswani2017attention}$^\dagger$    		& 0.82 & 0.88 & 0.70 & 1.22 & 0.80 & 0.88 \\
Ours   AP                                           				& 0.72  & 0.84 &\textbf{0.60} & 1.13 &\textbf{0.57} &  \textbf{0.77}\\
\hline
\hline
\end{tabular}
\end{center}
\caption{Results on the test partition of BP4D dataset  (in MSE values) $^\dagger$ denotes in-house evaluation} 
\label{Tab:bp4d_benchmarks_mae}
\end{table}

\end{document}